\documentclass[sigconf]{acmart}
\AtBeginDocument{%
  }

\usepackage{threeparttable}
\usepackage{balance}

\copyrightyear{2025}
\acmYear{2025}
\setcopyright{rightsretained}
\acmConference[MRAC '25] {Proceedings of the 3rd International Workshop on Multimodal and Responsible Affective Computing}{October 27--31, 2025}{Dublin, Ireland.}
\acmBooktitle{Proceedings of the 3rd International Workshop on Multimodal and Responsible Affective Computing (MRAC '25), October 27--31, 2025, Dublin, Ireland}
\acmISBN{979-8-4007-2052-9/2025/10}
\acmDOI{10.1145/3746270.3760215}

\begin{document}

\title{More Is Better: A MoE-Based Emotion Recognition Framework with Human Preference Alignment}

\author{Jun Xie}
\authornote{Both authors contributed equally to this research.}
\email{xiejun@lenovo.com}
\affiliation{%
  \institution{Lenovo Research}
  \city{Beijing}
  \country{China}
}

\author{Yingjian Zhu}
\authornotemark[1]
\email{zhuyingjian2024@ia.ac.cn}
\affiliation{%
  \institution{Institute of Automation, Chinese Academy of Sciences}
  \city{Beijing}
  \country{China}
}

\author{Feng Chen}
\email{chenfeng13@lenovo.com}
\affiliation{%
  \institution{Lenovo Research}
  \city{Beijing}
  \country{China}
}

\author{Zhenghao Zhang}
\email{zhenghaozhang2@outlook.com}
\affiliation{%
  \institution{University of Chinese Academy of Sciences}
  \city{Beijing}
  \country{China}}

\author{Xiaohui Fan}
\email{fxh23@mails.tsinghua.edu.cn}
\affiliation{%
  \institution{Tsinghua University}
  \city{Beijing}
  \country{China}
}

\author{Hongzhu Yi}
\email{yihongzhu23@mails.ucas.ac.cn}
\affiliation{%
 \institution{University of Chinese Academy of Sciences}
 \city{Beijing}
 \country{China}}

\author{Xinming Wang}
\email{wangxinming2024@ia.ac.cn}
\affiliation{%
  \institution{Institute of Automation, Chinese Academy of Sciences}
 \city{Beijing}
 \country{China}}

\author{Chen Yu}
\email{23125283@bjtu.edu.cn}
\affiliation{%
  \institution{Beijing Jiaotong University}
 \city{Beijing}
 \country{China}}

\author{Yue Bi}
\email{yuebi1207@gmail.com}
\affiliation{%
  \institution{Shandong University}
 \city{Beijing}
 \country{China}}

\author{Zhaoran Zhao}
\email{zhaozr3@lenovo.com}
\affiliation{%
  \institution{Lenovo Research}
 \city{Beijing}
 \country{China}}
 
\author{Xiongjun Guan}
\authornote{Corresponding author.} 
\email{gxj21@mails.tsinghua.edu.cn}
\affiliation{%
  \institution{Tsinghua University}
  \city{Beijing}
  \country{China}
}

\author{Zhepeng Wang}
\email{wangzpb@lenovo.com}
\authornotemark[2]
\affiliation{%
  \institution{Lenovo Research}
  \city{Beijing}
  \country{China}
}

\renewcommand{\shortauthors}{Jun Xie et al.}

\begin{abstract}
In this paper, we present our solution for the semi-supervised learning track (MER-SEMI) in MER2025. We propose a comprehensive framework, grounded in the principle that "more is better," to construct a robust Mixture of Experts (MoE) emotion recognition system. Our approach integrates a diverse range of input modalities as independent experts, including novel signals such as knowledge from large Vision-Language Models (VLMs) and temporal Action Unit (AU) information. To effectively utilize unlabeled data, we introduce a consensus-based pseudo-labeling strategy, generating high-quality labels from the agreement between a baseline model and Gemini, which are then used in a two-stage training paradigm. Finally, we employ a multi-expert voting ensemble combined with a rule-based re-ranking process to correct prediction bias and better align the outputs with human preferences. Evaluated on the MER2025-SEMI challenge dataset, our method achieves an F1-score of 0.8772 on the test set, ranking 2nd in the track. Our code is available at \href{https://github.com/zhuyjan/MER2025-MRAC25}{https://github.com/zhuyjan/MER2025-MRAC25}.
\end{abstract}




\maketitle

\section{Introduction}
\begin{figure*}[t]
\centering
\includegraphics[width=1.\textwidth]{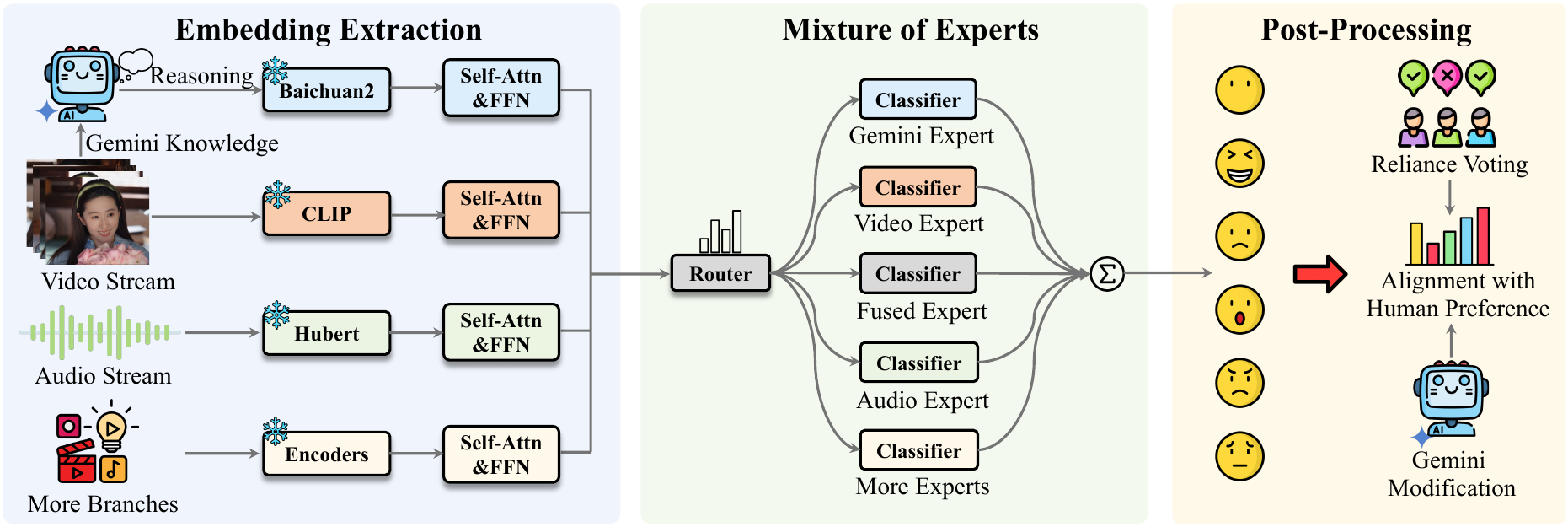}
\caption{The overall architecture of our proposed multimodal sentiment analysis framework. The model consists of three main stages: (1) \textbf{Embedding Extraction}, where features from diverse modalities (e.g., video, audio, reasoning) are extracted with powerful encoders like CLIP, Hubert, and Baichuan2. (2) \textbf{Mixture of Experts (MoE)}, where a router module dynamically weights the predictions from various unimodal and fused experts. (3) \textbf{Post-Processing}, where the aggregated output is further refined and aligned with human preferences through reliance voting and VLM-based modification.}
\label{fig1}
\end{figure*}
Emotion is fundamental to human interaction and the development of empathetic human-computer interaction (HCI) systems~\cite{hci1, hci2}. Multimodal Emotion Recognition (MER) \cite{mer23, mer24, merbench}, a core area within affective computing, aims to interpret human emotional states by integrating multiple data channels, such as facial expressions, vocal intonation, and textual content. This research offers significant academic value and has substantial application potential in fields like smart driving \cite{car}, healthcare \cite{depression}, and education \cite{education}. By enabling machines to understand and adapt to user emotions, MER facilitates more harmonious and effective human-machine collaboration.

The MER2025-SEMI Challenge~\cite{mer25} provides a limited amount of labeled training data and encourages participants to leverage large volumes of unlabeled data to improve model performance. Most previous studies have focused on selecting more effective encoders~\cite{emotionllama, qi2024multimodal, yu2024emotion, zhao2024improving} and designing complex fusion mechanisms~\cite{cl, shi2024audio, ge2024early}. For instance, Qi et al.~\cite{qi2024multimodal} adopted a prompt-learning approach to fine-tune CLIP, resulting in a more robust video encoder. Shi et al.~\cite{shi2024audio} proposed an Audio-Guided Fusion strategy to integrate features from multiple modalities. However, excessive model complexity often leads to overfitting. Therefore, instead of designing overly complex fusion mechanisms or fine-tuning encoders, we adopt a simple multi-expert voting approach. This strategy enables multiple expert models to collaborate and make collective decisions, compensating for the limitations of individual models by leveraging the diversity and breadth of knowledge across the ensemble.

Existing pseudo-labeling strategies for utilizing unlabeled data exhibit significant limitations. These methods~\cite{qi2024multimodal, cl} typically depend on the predictions of pre-trained models to generate supplementary training data, thereby forming a self-reinforcing loop. This often results in "confirmation bias," where the model keeps reinforcing its existing beliefs and struggles to learn new patterns. To break this cycle of knowledge closure, we introduce Gemini~\cite{gemini}, a large vision-language model (VLM) with stronger generalization capabilities, to validate the model’s predictions, thereby significantly improving the quality of pseudo-labels.

Furthermore, annotator bias is a prevalent issue in emotion recognition. Due to individual differences and the inherent subjectivity of emotional perception, the GT labels provided by different annotators often contain bias. Consequently, models trained on such data inevitably exhibit systematic deviations from the ideal judgment. To address this issue, we acknowledge that relying solely on the model’s learning capabilities is insufficient. Therefore, we propose a post-processing alignment step following the model’s prediction.

Our main contribution can be summarized as follows:
\begin{itemize}
    \item Rather than fixating on the choice of input modality, or striving to identify the most optimal encoder, we adopt the principle of "more is better." We advocate for integrating as many input branches as possible, treating each branch as an independent expert, ultimately forming a powerful Mixture of Experts (MoE) system.
    \item To address the limitations of existing pseudo-labeling strategies, we propose a novel consensus-based pseudo-label generation method, along with a complementary two-stage training paradigm. This approach allows for more effective utilization of large-scale unlabeled data.
    \item We introduce a post-processing strategy aimed at aligning predictions with human preferences. By calibrating the model’s output distribution with the class distribution of the test dataset, we significantly enhance the final prediction accuracy.
\end{itemize}

\section{Method}

\subsection{More Signals, Broader Perspective}
\begin{figure*}[t]
\centering
\includegraphics[width=1.\textwidth]{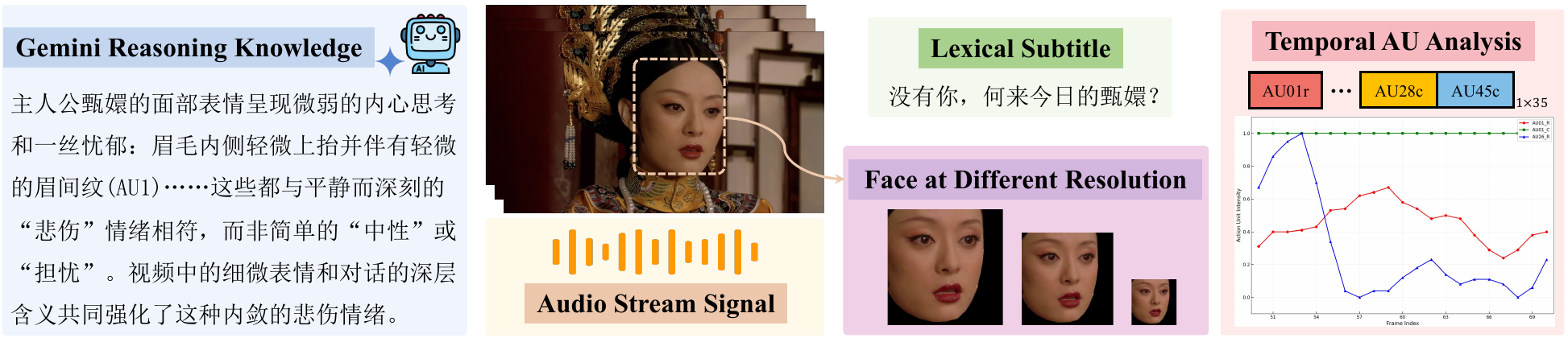}
\caption{More signal examples in our method.}
\label{fig2}
\end{figure*}

As illustrated in Fig. \ref{fig1}, the core design of our method involves integrating as many sentiment analysis experts as possible to form a Mixture of Experts (MoE) system, which leverages their diverse knowledge perspectives to collaboratively produce the final prediction. Specifically, our model can integrate up to 18 different experts, and the "more branches" represent additional signals such as caption text, facial features, action unit analysis~\cite{au1, au2}, and others. The knowledge of each expert is extracted through their corresponding encoders~\cite{clip, hubert, baichuan2}, followed by self-attention and FFN. The Router then assigns weight scores to each expert by concatenating the features from all preceding branches and passing the combined representation through a linear layer. The Fused Expert is generated by concatenating all the features from the previous branches into a unified representation. The outputs of all experts are aggregated through a weighted sum to produce the emotion label. We experimented with various combinations of modalities and expert quantities. For the combinations achieving higher scores, their predictions are aligned with human judgment preferences through the post-processing.

Our final architecture comprises six distinct input branches: full-frame images, cropped face images, caption text, audio stream, Gemini knowledge, and temporal AU (Action Unit) analysis. In contrast to the MER2025 baseline~\cite{mer25}, which uses cropped facial images as visual input, the full-frame input includes entire video frames along with background information. Full-frame images provide richer contextual cues that potentially convey more comprehensive information about the individual than isolated facial features extracted through OpenFace~\cite{openface}. Additionally, our experiments reveal that face images of varying resolutions exhibit significant performance differences in unimodal evaluations. Therefore, we treat each resolution as a distinct feature branch. In addition to the modalities commonly used in previous works, we also propose two novel feature branches, described as follows.

\subsubsection{Gemini Knowledge}
As illustrated in the blue part of Fig. \ref{fig2}, we use a carefully designed prompt (refer to our \href{https://github.com/zhuyjan/MER2025-MRAC25}{code}) to input raw videos with audio tracks into Gemini-2.5-Flash for emotion analysis. The model outputs its response in JSON format, which contains the following fields: (1) Reasoning Process: A detailed analysis of the main character’s emotions based on facial expressions (action units), body posture, head pose, inter-frame relationships, audio features (pitch, speech rate, volume, and tone quality), and other relevant indicators. (2) Confidence Scores: Confidence levels for six emotions—neutral, angry, happy, sad, worried, and surprised. (3) Label: For ease of data processing, we instruct Gemini to output a final predicted emotion label. (4) Modality Contribution: The predictive influence of each modality is dynamic. For instance, audio cues might dominate in one clip, while visual information is more critical in another. To leverage this, we prompt Gemini to output its degree of reliance on each modality, enabling it to better weigh their specific insights for the final prediction. As shown in Fig. \ref{fig1}, the resulting JSON file for each video clip is treated as a new signal branch and is fed into the Baichuan2-7B text encoder to extract textual features.

\subsubsection{Temporal AU Analysis}
Variations in action units (AUs) are crucial for analyzing human emotions. In previous studies, facial information has primarily been utilized by extracting frame-level visual features through visual encoders, which are then aggregated using global average pooling to obtain an utterance-level facial representation. While this approach has achieved reasonable performance, it overlooks important temporal dynamics that are critical in emotion recognition. To address this limitation, we propose temporal AU analysis as a novel signal branch.

As illustrated in the red region of Fig. \ref{fig2}, the Action Unit (AU) information for each frame is encoded as a 35-dimensional vector, corresponding to the 35 defined facial AU components. This yields an initial input tensor with a shape of $B \times T \times 35$, where $B$ denotes the batch size, $T$ is the number of frames. Subsequently, a linear layer projects these 35-dimensional vectors into a d-dimensional feature space, transforming the tensor's shape to $B \times T \times d$. To aggregate temporal features, a learnable [CLS] token with dimensions $B \times 1 \times d$ is prepended to the sequence, modifying the input shape to $B \times (T+1) \times d$. The sequence is then processed using self-attention and a feed-forward network (FFN), which yields an output of the same dimensions. Finally, the representation corresponding to the [CLS] token is extracted to serve as the temporal AU feature.

\subsection{More Samples, Fuller Distribution} \label{subsec:data}
A primary challenge in MER2025-SEMI is the scarcity of labeled training data, which often leads to overfitting, especially with complex model architectures. A common strategy to address this is pseudo-labeling, where a model's own predictions on unlabeled data are used as new training targets. However, this approach has a fundamental limitation: it essentially confines the model to its learned data distribution and can amplify its inherent biases. To overcome this limitation and effectively leverage the unlabeled dataset, we introduce a novel two-stage training paradigm centered on a consensus-based pseudo-labeling strategy. 

\subsubsection{Pseudo Labeling}
First, we train an efficient baseline model purely on the provided labeled set. This model, which represents an optimal trade-off between performance and complexity, is then used to generate predictions for the entire unlabeled test set. Concurrently, we leverage the powerful capabilities of the Gemini to obtain predictions for the same data.  The core of our strategy lies in the subsequent filtering step: we create a high-confidence pseudo-labeled dataset by selecting only those samples where the predictions from our baseline model and Gemini are identical. This consensus between a task-specific model and a large VLM acts as a powerful filter, significantly increasing the quality and reliability of the labels.

\subsubsection{Two-stage Training}
Although the generated pseudo-labeled data are of high quality, they may still exhibit distributional bias relative to the given labeled data. To mitigate the potential impact of this bias on the labeled data, we adopt a two-stage training paradigm. First, we pretrain the model on the pseudo-labeled data, enabling it to learn more robust and generalizable feature representations from a larger and more diverse dataset. Then, we fine-tune the model on the labeled training data using a learning rate that is 10 times smaller than that used during pretraining.

\subsection{More Deliberation, Less Bias} \label{subsec:bias}
To overcome the limitations inherent in a single model, we introduced a post-processing strategy aimed at achieving more robust outcomes. 
This approach effectively reduces the influence of random biases introduced by individual models.
By averaging or weighting their predictions, the system achieves more robust and consistent outputs, even in the presence of noisy or ambiguous data \cite{xie2025four}. 
Moreover, this strategy enhances the generalization ability of the overall framework, as the diverse decision patterns of the experts can better adapt to heterogeneous or unseen samples.
Specifically, we first determined the individual prediction accuracy of each expert and utilized this as a confidence measure to perform weighted voting on the aggregated results.
Let ${\hat{y}_i}$ and ${y_i}$ represent the prediction results and the actual emotions, respectively, the reliability $r$ of a certain expert $e$ is calculated as
\begin{equation}
    r_{e} = \sum_i \mathbb{I} (y_i = \hat{y}_i^e ) \:/ \: n \;,
    \label{eq:weight}
\end{equation}
where $\mathbb{I}$ is the indicator function, $n$ is total number of samples evaluated.
The final result is classified based on
\begin{equation}
    \hat{y} = \operatorname{argmax}_{k} \sum_{e=1}^{M} r_{e} \cdot p_{k}^e\;,
    \label{eq:vote}
\end{equation}
where $M$ is the total number of experts involved in decision-making,  $p_k$ is the probability of the $k$-th emotion.

Notably, Equation \ref{eq:vote} yielded significantly improved results without training, yet we observed a strong bias towards predicting ‘neutral’.
A plausible explanation is that neutral resides at the ‘center’ of the emotional spectrum, making it inherently more susceptible to confusion and misjudgment compared to other, more distinct emotional states.
On the other hand, we observed that the closed-source VLM demonstrated better performance on emotions with relatively clearer decision boundaries, such as ‘angry’, ‘happy’ and ‘surprise’. 
However, it struggled to differentiate between ‘neutral’ and other emotions, as well as between ‘worried’ and ‘sad’.
Therefore, we further introduced rule-based adjustments to re-rank the results.
Based on the confusion matrix of emotion classification and the category ratio of predicted results and ground truth, the final rule adopted is as follows:
\begin{itemize}
    \item When the highest number of votes is ‘neutral’ and the second highest is ‘angry’, modify it to ‘angry’.
    \item When the maximum number of votes is ‘neutral’, and the second highest number of emotional votes exceeds 1/4, it is corrected to that category.
    \item When the maximum number of votes is ‘neutral’ and VLM considers it to be ‘angry’, ‘happy’ or ‘surprise’, modify it to the corresponding emotion.
\end{itemize}
The comparison of emotional distribution before and after adjustment is shown in Table \ref{tab:emo_distribution}.
It can be seen that the core of these strategies lies in carefully refining the decision boundaries that distinguish neutral from other emotions. 
This pivotal step seeks to align the model’s categorizations with the subtle preferences of human experts.

\begin{table}[!t]
  \centering
  \begin{threeparttable} 
    \caption{The distribution of emotions under different strategies.
    From top to bottom, they are: ground truth of validation set, the single model prediction results / the voting results /  the re-ranked results of the test set.}
    \label{tab:emo_distribution}
    \small
    \begin{tabular}{c|cccccc}
      \toprule
      {} & \textbf{neu} & \textbf{ang} & \textbf{hap} & \textbf{sad} & \textbf{wor} & \textbf{sur}\\
      \midrule
        {valid} & 24.8\% & 24.0\% & 20.6\% & 14.5\% & 12.2\% & 3.8\% \\
        \hline
        {single} & 48.4\% & 14.2\% & 10.2\% & 13.3\% & 11.0\% & 2.9\% \\
        {vote} & 49.2\% & 14.9\% & 10.7\% & 13.6\% & 8.8\% & 2.9\% \\
        {re-rank} & 41.6\% & 18.7\% & 11.5\% & 15.6\% & 9.7\% & 3.0\% \\
      \bottomrule
    \end{tabular}
  \end{threeparttable}
\end{table}

\section{Experiments}

\subsection{Datasets}
MER2025-SEMI provides two sets, Train \& Val with labels and Test without labels, containing 7,396 and 124,802 videos respectively.
The final test set contains 20,000 videos.
Since the Train \& Val dataset is not split into separate training and validation sets, similar to the baseline system, we adopt a five-fold cross-validation approach. 
The best results from each of the five validation sets are then equally weighted to calculate the final result. 

\subsection{Implementation Details}
All training processes are performed with an initial learning rate of 1e-3 (end of 1e-4), cosine annealing scheduler, default AdamW optimizer and batch size of 256 until convergence (about 10 epochs).
During the preparation phase of multi-stage training, we first train an initial model using the Train \& Val dataset. 
Next, we generate pseudo labels for 20,000 samples and train a new model with these pseudo labels to equip it with foundational emotion recognition capabilities. 
Finally, we fine-tune the model using high-quality labels from the Train \& Val dataset to correct the bias and enhance its performance.
By selecting different branch features and dividing the training sets in various ways, we ultimately developed 15 distinct models. 
Their performance was further enhanced through the ensemble strategy and reordering rules described in Section \ref{subsec:bias}.

\subsection{Results}
In experiments, the impact of single-modal encoder was first evaluated and then fixed. 
Table \ref{tab:single_modal} provides a relative comparison of leading models for reference.
On this basis, we make incremental adjustments to the model structure, loss function, and voting strategy. 
The key ablation experiment results are presented in Table \ref{tab:ablation}.
The second group shows that adjusting the model structure and loss function improved the baseline by 1\%. 
This improvement is attributed to two factors: (1) the finer-grained attention interaction mechanism of the transformer, and (2) the loss function’s role in balancing emotion categories with varying proportions and difficulty levels.
Furthermore, incorporating new input signals has led to remarkable improvements. 
One possible explanation is that using frozen parameters in the encoder may result in the loss of emotion related representations. 
The additional input features effectively supplement valuable missing information.
Finally, with the support of pre-training and model integration, the accuracy of our solution has been significantly enhanced once again.
This strongly validates our analysis in Sections \ref{subsec:data} and \ref{subsec:bias}.

\begin{table}[!t]
  \centering
  \begin{threeparttable} 
    
    \caption{Top 3 Models and corresponding fscores in single modal benchmark testing.
    The reported score is the average of 5-fold cross validation of Train \& Val dataset}
    \label{tab:single_modal}
    \footnotesize
    \begin{tabular}{cc|cc|cc}
      \toprule
       \multicolumn{2}{c|}{{\textbf{Visual}}} 
       & \multicolumn{2}{c|}{{\textbf{Audio}}} 
       & \multicolumn{2}{c}{{\textbf{Text}}} \\
       \cmidrule(lr){1-2}\cmidrule(lr){3-4}\cmidrule(lr){5-6}
      \textbf{Method} & \textbf{fscore} & \textbf{Method} & \textbf{fscore} & \textbf{Method} & \textbf{fscore}\\
      \midrule
         clip-vit~\cite{clip} & 0.67 & chn-hubert~\cite{hubert} & 0.75 & Baichuan2~\cite{baichuan2} & 0.56 \\
         fg-clip~\cite{fgclip} & 0.65 & Qwen2-Audio~\cite{qwen} & 0.68 & bloom~\cite{bloom} & 0.53 \\
         InternVL~\cite{InternVL} & 0.62 & chn-wav2vec2~\cite{chnw} & 0.67 & chn-roberta~\cite{chnr} & 0.53 \\
      \bottomrule
    \end{tabular}
  \end{threeparttable}
\end{table}

\begin{table}[!t]
  \centering
  \begin{threeparttable} 
    \caption{Ablation study of our approach. The reported score comes from the official evaluation results of the competition.}
    \label{tab:ablation}
    \small
    \begin{tabular}{cccccc|c}
      \toprule
      \textbf{Model} & \textbf{Loss} & \textbf{feat} & \textbf{pretrain} & \textbf{Vote} & \textbf{Re-rank} & \textbf{fscore} \\
      \midrule
      baseline & ce & 3 & \scriptsize\textbackslash& \scriptsize\textbackslash & \scriptsize\textbackslash & 0.7580 \\
      \hline
      ours & ce & 3 & \scriptsize\textbackslash& \scriptsize\textbackslash & \scriptsize\textbackslash & 0.7673 \\
      ours & focal~\cite{focal} & 3 & \scriptsize\textbackslash& \scriptsize\textbackslash & \scriptsize\textbackslash & 0.7676 \\
      ours & LMF~\cite{lmf} & 3 & \scriptsize\textbackslash& \scriptsize\textbackslash & \scriptsize\textbackslash & 0.7684 \\
      \hline
      ours & LMF~\cite{lmf} & 6 & \scriptsize\textbackslash & \scriptsize\textbackslash & \scriptsize\textbackslash & 0.8430 \\
      \hline
      ours & LMF~\cite{lmf} & 6 & \checkmark & \scriptsize\textbackslash & \scriptsize\textbackslash & 0.8547 \\
      \hline
      ours & LMF~\cite{lmf} & 6 & \checkmark & \checkmark & \scriptsize\textbackslash & 0.8655 \\
      ours & LMF~\cite{lmf} & 6 & \checkmark & \checkmark & \checkmark & 0.8772 \\
      \bottomrule
    \end{tabular}
  \end{threeparttable}
\end{table}

\section{Conclusion}
In this paper, we presented a Mixture of Experts (MoE) system for multimodal emotion recognition that integrates a wide array of input modalities, including novel signals like VLM-generated knowledge and temporal AU analysis. To effectively leverage the vast amount of unlabeled data, we introduced a consensus-based pseudo-labeling strategy coupled with a two-stage training paradigm, significantly enhancing the model's robustness and generalization. Finally, we addressed inherent model and annotator biases through a multi-stage post-processing pipeline. Extensive experiments validate the effectiveness of each component, culminating in a final F-score of 0.8772, ranking 2nd in the MER-SEMI 2025 track.

\clearpage


\balance
\bibliographystyle{ACM-Reference-Format}
\bibliography{sample-base}

\appendix

\end{document}